\documentclass[10pt,twocolumn]{ICCAS2024}
 
\usepackage{diagbox}
\usepackage{graphicx}
\usepackage{float}
\begin{document}

\title{Low Clearance Hinge Joint Mechanism Based on 3D Printing on Sheet Fabrication Methodology}

\author{Jaehyung Jang${}^{1}$, Euibin Shin${}^{1}$, Allison M. Okamura${}^{2}$, and Jee-Hwan Ryu${}^{1*}$ }

\affils{ ${}^{1}$Department of Civil and Environmental Engineering, KAIST, \\
Daejeon, 34141, Korea (jhryu@kaist.ac.kr) {\small${}^{*}$ Corresponding author}\\
${}^{2}$Department of Mechanical Engineering, Stanford University, \\
Stanford, CA 94305, USA (aokamura@stanford.edu)}


\abstract{
This paper presents a low clearance hinge joint mechanism based on the 3D printing on sheet fabrication method. This approach simplifies the manufacture of hinge mechanisms and overcomes the limitations of conventional origami manufacturing by eliminating the need for adhesives commonly used in assembly, making it suitable for use in robots at the tens of centimeter scale. The advantages and disadvantages of three types of hinge joint mechanisms are presented, and the hinge joint, capable of being designed with low clearance for various facet thickness, was selected. Based on the selected hinge joint, the twisting angle and bending force of the hinge joint are analyzed, leading to the implementation of a 0.1mm clearance in the application. Torsional resistance was tested to measure the torque required for twisting due to plastic deformation and clearance, respectively. The results indicated that the torque required for plastic deformation is sufficient to constrain the hinge joint’s degrees of freedom, while the torque required for twisting due to clearance is minimal. The proposed hinge joint mechanism applied 3 degrees of freedom delta robot manipulator developed from the analyzed data, and it provided precise motion with low clearance.
}


\maketitle


\section{Introduction}

Over the past few decades, pin joints have been applied in various systems including automation equipment and robots due to their precise movement and simple kinematics. However, the complexity of the components and the assembly increased the manufacturing time and cost of the systems. Origami technology, which leverages the principles of paper folding to create complex 3D structures from simple 2D forms, has been employed to simplify manufacturing and reduce costs by using flexible sheet-based hinge joints instead of pin joints \cite{niiyama2015pouch}, \cite{lee2017origami}, \cite{zhakypov2019programmable} and \cite{zuliani2021variable}. In addition, its compactness, light weight, and inherent compliance have made systems safer and more robust in dynamic environments, and have excelled in the development of miniature robots that were previously difficult to manufacture \cite{mcclintock2018millidelta}, \cite{mintchev2019portable}, and \cite{williams20224}.

Although their remarkable advantages have brought revolutionary changes to robotic design, origami's adhesive-based assembly faces limitations in practical applications. The drawbacks of adhesives, such as significantly reduced adhesion when exposed to humid or high-temperature environments, and a relatively short life span, restrict their applicability. To address these issues, 3D printing on fabric technology has been used. In this technology, porous fabric is inserted during the 3D printing process, allowing the top and bottom facets of the origami structure to bond together without the use of adhesives. Based on 3D printing on fabric technology, \cite{choi2023fabrication} presents an origami gripper that overcomes the disadvantages of adhesives. Additionally, the intrinsic characteristics of 3D printing allow for the integration of complex-shaped facets with creases. However, this approach has its own limitations. Due to the low stiffness of the fabric, it lacks torsional resistance, and the high clearance prevents its application in systems that require high precision.

In this study, we introduce a 3D printing on sheet-based low clearance hinge joint mechanism that allows for the assembly of origami structures without adhesives. Since the proposed mechanism is fabricated using a 3D printer, it ensures easy manufacturing and assembly, as well as customizable production, thereby reducing both manufacturing time and cost. It also works effectively in humid or high-temperature environments without performance degradation. In addition, it provides high-precision rotational movement for automation systems. We applied this joint mechanism to a 3 degree-of-freedom (DoF) delta robot manipulator requiring high precision on the scale of tens of centimeters and validated the mechanisms' performance.

\section{Fabrication methodology}

\begin{figure}[t]
    \centering
    \includegraphics[width=.9\linewidth]{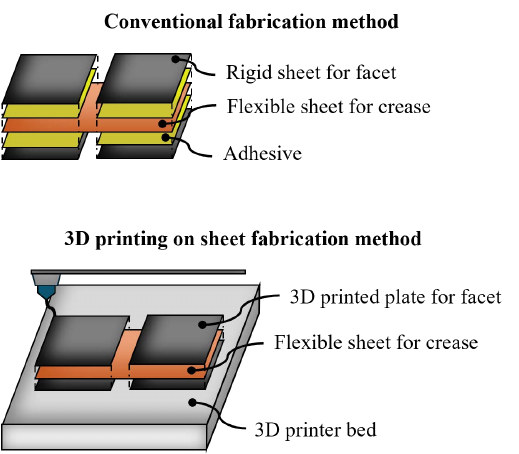}
    \caption{Fabrication method for hinge joint mechanism.}
    \label{fig:fabrication}
\end{figure}

Origami structure have been applied to various robots due to their inherent compliance, lightweight and compact mechanisms, and flat-foldable characteristics \cite{russo2016soft}, \cite{faber2018bioinspired}, \cite{kim2019bioinspired}, and \cite{lee2021high}. Notably, unlike traditional robots, they can be fabricated using industrially compatible methods, addressing issues related to the miniaturization and mass production of conventional robots . A commonly used fabrication method for origami structures is laser cutting. Origami structures consist of facets and creases, and an adhesive is required to bond these two components together. Each of these three components is individually produced by laser cutting, and the origami structure is assembled by layering all the components together, as shown in the top of Fig.\ref{fig:fabrication}. However, adhesive based assembly has several drawbacks due to the inherent limitations of adhesives. For example, adhesives generally have lower adhesion strength than mechanical bonding, and their performance can degrade over time due to aging, heat, and humidity. Additionally, adhesives are susceptible to significant decreases in adhesion strength when exposed to high temperatures and liquids.

Therefore, this paper proposes a fabrication method for origami structure without using adhesives by employing 3D printing on sheet technology as shown in the bottom of Fig.\ref{fig:fabrication}. The principle of 3D printing on sheet is straightforward. First, the bottom facet is printed using a 3D printer, which is then paused. Next, a laser-cut flexible sheet is placed on top of the printed facet. The 3D printer is then resumed to print the top facet. In this way, the flexible sheet is inserted between the facets and held in place without adhesives. This approach minimizes manual assembly work and overcomes the problems associated with traditional adhesive based assembly. 

\subsection{Selection of hinge joint mechanism}

\begin{figure}[t]
\begin{center}
\includegraphics[width=.7\linewidth]{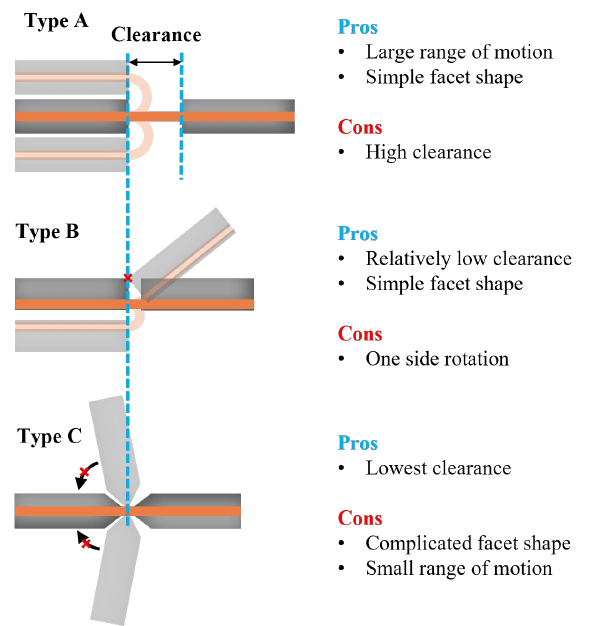}
\caption{Pros and cons of three different types of hinge joints.}
\label{fig:joint_type}
\end{center}
\end{figure}

Clearance is an important factor that determines the accuracy and precision of rotational movement in joint. Therefore, in this chapter, we describe the types of hinge joint mechanisms, their pros and cons, and explain which type is suitable for low-clearance hinge joint based on 3D printing on sheet fabrication method. Fig.\ref{fig:joint_type} shows three types of hinge joint mechanisms. Type A is the most commonly used hinge joint mechanism, featuring a large range of motion and a simple facet shape. Therefore, it is suitable for complex origami structures. However, this type has a drawback: as the thickness of the facet increases, the clearance also significantly increases, making it difficult to adapt to systems larger than tens of centimeters. Type B has facets of different thicknesses on both sides, resulting in low clearance while allowing the use of thicker facets. However, this type only allows for one-sided rotation, and due to the nature of 3D printing, it suffers from reduced adhesion strength in thin facets. 

Lastly, type C, although it can achieve the lowest clearance among hinge joint mechanisms, has the drawback of a more complex facet shape and a limited range of motion. However, 3D printing on sheet technology allows for easy printing of complex facet shapes, and the range of motion provided by type C is sufficient for implementing a 3 DoF delta robot manipulator. Therefore, in this study, type C was employed to develop a low-cost, customizable 3 DoF delta robot manipulator with high precision and accuracy through low clearance, and is robust in humid and high-temperature environments.

\subsection{Analysis of clearance}

\begin{figure}[t]
\begin{center}
\includegraphics[width=.7\linewidth]{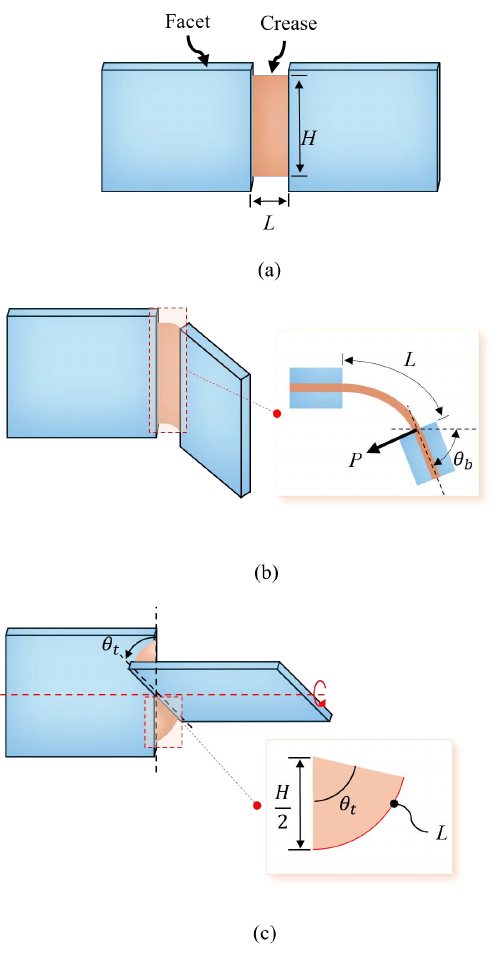}
\caption{ Geometry of hinge joint mechanisms in (a) normal state (b) bending, and (c) twisting.}
\label{fig:Clearance}
\end{center}
\end{figure}

\begin{figure}[t]
\begin{center}
\includegraphics[width=.7\linewidth]{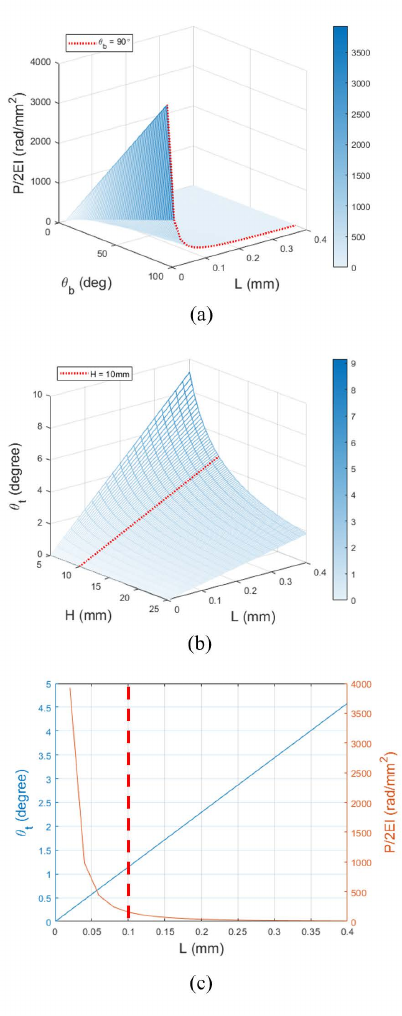}
\caption{(a) Twisting angle of hinge joint due to clearance (b) Bending force with respect to $\theta_b$ and $L$ (c) twisting angle and bending force at $H$ = 10mm and $\theta_b$ = 90$^{\circ}$.}
\label{fig:analysis}
\end{center}
\end{figure}

The hinge joint, as illustrated in Fig.\ref{fig:Clearance} (a), consists of facets and a crease, with the distance between them defined as clearance. To select the appropriate clearance, which affects the performance of the hinge joint, the bending force ($P$) and twisting angle ($\theta_t$) of the hinge caused by clearance are considered. As shown in Fig.\ref{fig:Clearance} (b), when a beam deforms by $P$, the maximum slope ($\theta_b$) of hinge joint is calculated by Euler-Bernoulli beam theory \cite{craig2020mechanics} as follows:

\begin{eqnarray}
\begin{aligned}
\theta_b = \frac{PL^2}{2EI}
\label{eq:bending angle}
\end{aligned}
\end{eqnarray}

Where $L$ is the crease length (clearance), $E$ is the modulus of elasticity, and $I$ is the second moment of inertia of the crease's cross-section. By rearranging this equation to solve for the force, it can be expressed as follows:

\begin{eqnarray}
\begin{aligned}
P = 2EI\frac{\theta_b}{L^2}
\label{eq:bending angle force form}
\end{aligned}
\end{eqnarray}

Based on the equations, Fig.\ref{fig:Clearance} illustrates the twisting that occurs in the hinge joint and the corresponding angle. The twisting angle ($\theta_t$) is defined by the height ($H$) and length ($L$) of the crease as follows:

\begin{eqnarray}
\begin{aligned}
\theta_t = \frac{2L}{H}
\label{eq:twisting angle}
\end{aligned}
\end{eqnarray}

Fig.\ref{fig:analysis} demonstrates the relationships between $P$ and $\theta_t$. Fig.\ref{fig:analysis} (a) shows that as the $L$ decreases, $P$ increases dramatically for all values of $\theta_b$. This indicates that shorter crease lengths demand significantly higher forces to achieve the same amount of bending. As shown in Fig.\ref{fig:analysis} (b), as the clearance decreases, the $\theta_t$ also decreases. In addition, for the same $L$, increasing $H$ results in a smaller $\theta_t$. In this study, considering the size of the 3 DoF delta robot manipulator, we designed the hinge joint with a $H$ of 10mm. Given the characteristics of the type C hinge joint, it allows a maximum range of motion up to 90 degrees in one direction. Therefore, taking these two conditions into account, the relationship between $L$, $P$, and $\theta_t$ is depicted in Fig.\ref{fig:analysis} (c). Although $\theta_t$ is minimized when $L$ is 0, which would theoretically provide the highest accuracy and precision, $P$ becomes exponentially large at the point. Therefore, considering practical constraints, we selected a design value of $L$ at 0.1 mm.

\subsection{Torsional resistance test of hinge joint}

\begin{figure}[t]
\begin{center}
\includegraphics[width=.6\linewidth]{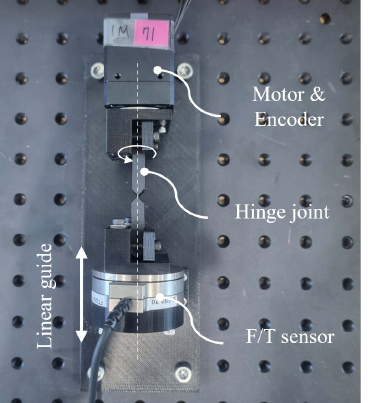}
\caption{Test setup to measure torsional resistance of hinge joint.}
\label{fig:test setup}
\end{center}
\end{figure}

The hinge joint samples were tested to measure the hinge joint's torsional resistance. The tests were conducted on two hinge joints with clearances of 0 mm and 0.4mm, respectively. As shown in Fig.\ref{fig:test setup}, the test setup consists of a servo motor (Dynamixel XH540-W150-R, ROBOTICS) equipped with an encoder to rotate the hinge joint at a constant speed, a 6-axis F/T sensor (mini 45, ATI industrial automation) to measure the torsional resistance, and a linear guide.

\begin{figure}[t]
\begin{center}
\includegraphics[width=.7\linewidth]{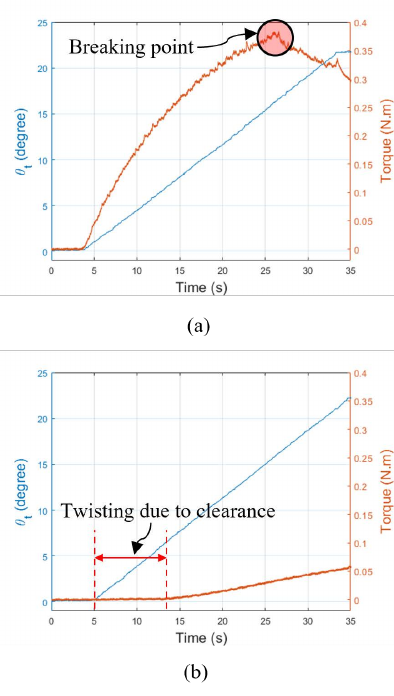}
\caption{The experimental results of torsional resistance test with (a) 0 mm clearance and (b) 0.4 mm clearance.}
\label{fig:experiments}
\end{center}
\end{figure}

Fig.\ref{fig:experiments} presents the experimental results. As shown in Fig.\ref{fig:experiments} (a), twisting occurs even without clearance when the applied torque is sufficiently high. This phenomenon is probably due to plastic deformation of the sheet when the torque exceeds a certain threshold. In addition, the torque required to twist the joint increases exponentially with the angle, and the reaction torque decreases after the breaking point.

In contrast, as illustrated in Fig.\ref{fig:experiments}, the hinge joint with a 0.4 mm clearance initially exhibits a relatively low increase in reaction torque initially due to twisting within the clearance. However, once the twisting exceeds the range caused by the clearance, the reaction torque increases sharply.

\section{Delta robot manipulator}

\subsection{Design of 3 DoF delta robot manipulator}

\begin{figure}[t]
\begin{center}
\includegraphics[width=.6\linewidth]{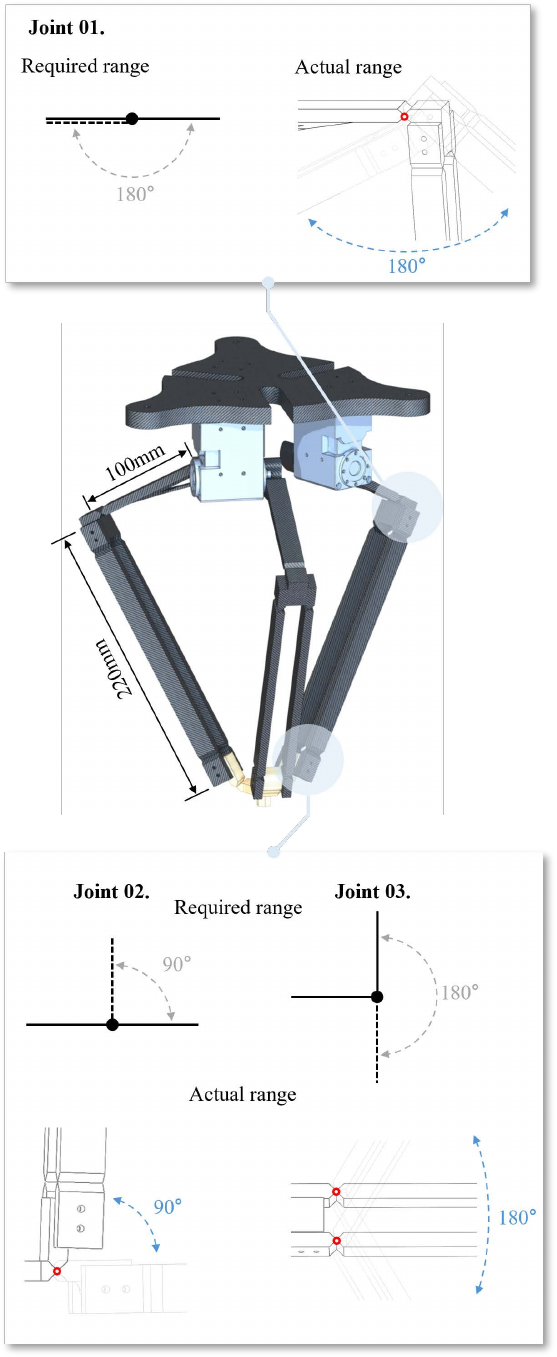}
\caption{Design of 3 DoF delta robot manipulator and its joints.}
\label{fig:design delta robot}
\end{center}
\end{figure}

Considering the design parameters' values determined in previous chapters, we designed and fabricated a 3 DoF delta robot manipulator on the scale of several tens of centimeters. As shown in Fig.\ref{fig:design delta robot}, the 3 DoF delta robot manipulator has three primary joints (joint 01, joint 02, and joint 03). The length of the lower link is 220 mm, and the that of the upper link is 100 mm. The type C hinge joint provides a sufficient range of motion for joint 02 and joint 03. However, joint 01 requires a range of motion of 180 degrees in one direction, which the type C hinge joint cannot cover. Therefore, we assembled two links vertically to achieve a 180-degree range of motion in one direction, ensuring that joint 01 meets the required range of motion.

\subsection{Prototype of 3 DoF delta robot manipulator}

\begin{figure}[t]
\begin{center}
\includegraphics[width=.9\linewidth]{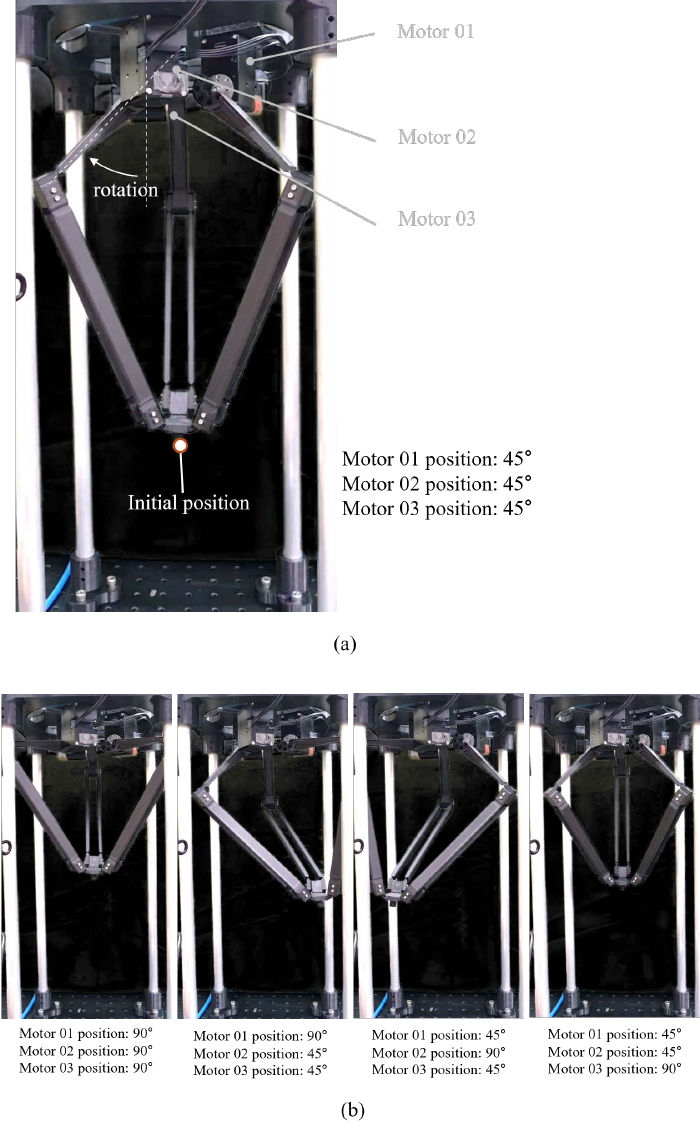}
\caption{Fabricated 3 DoF delta robot manipulator and its actuation}
\label{fig:fabricated delta robot}
\end{center}
\end{figure}

Fig.\ref{fig:fabricated delta robot} shows the fabricated 3 DoF delta robot manipulator and its actuation. The manipulaotr is equipped with three servo motors (Dynamixel XH540-W150-R, ROBOTICS) that provide rotational actuation. All the joints of the manipulator are based on hinge joint mechanisms fabricated using 3D printing on sheet technology, making them completely free of bearings and pins. The initial position of the manipulator is set with all motor angles at 45 degrees, and the motors are actuated up to a maximum range of 90 degrees. During the four different actuations, the manipulator demonstrates minimal tip rotation, indicating that the joint clearance is sufficiently small. This observation underscores the effectiveness of hinge joint design in maintaining precision and accuracy in the manipulator's movements.

\section{Conclusion}
This paper proposed a low-clearance hinge joint mechanism based on a 3D printing on sheet fabrication methodology, which provides precise and accurate movements due to its low-clearance design. This mechanism is applicable to robots on a tens of centimeter scale. Furthermore, the 3D printing on sheet fabrication methodology addresses the issues associated with traditional origami fabrication methods by eliminating the need for adhesives, which are typically used in assembly and often introduce problems. This approach presents potential for reducing manufacturing time and costs for robots that require high precision and accuracy.

However, our investigation revealed that the relationship between sheet thickness and factors such as plastic deformation, clearance, and bending force have not been adequately considered. Additionally, the effects of increased sheet thickness on the performance of 3D printing remain insufficiently explored. Therefore, future research should focus on comprehensively analyzing these relationships to optimize the design and functionality of hinge joints based on 3D printing on sheet fabrication.

\section*{ACKNOWLEDGEMENT}
This research was supported by the MOTIE (Ministry of Trade, Industry, and Energy) in Korea, under the Fostering Global Talents for Innovative Growth Program (P0017303) supervised by the Korea Institute for Advancement of Technology (KIAT)

\bibliographystyle{ieeetr}
\bibliography{references.bib}

@article{lee2017origami,
  title={Origami wheel transformer: A variable-diameter wheel drive robot using an origami structure},
  author={Lee, Dae-Young and Kim, Sa-Reum and Kim, Ji-Suk and Park, Jae-Jun and Cho, Kyu-Jin},
  journal={Soft robotics},
  volume={4},
  number={2},
  pages={163--180},
  year={2017},
  publisher={Mary Ann Liebert, Inc. 140 Huguenot Street, 3rd Floor New Rochelle, NY 10801 USA}
}

@article{niiyama2015pouch,
  title={Pouch motors: Printable soft actuators integrated with computational design},
  author={Niiyama, Ryuma and Sun, Xu and Sung, Cynthia and An, Byoungkwon and Rus, Daniela and Kim, Sangbae},
  journal={Soft Robotics},
  volume={2},
  number={2},
  pages={59--70},
  year={2015},
  publisher={Mary Ann Liebert, Inc. 140 Huguenot Street, 3rd Floor New Rochelle, NY 10801 USA}
}

@inproceedings{zuliani2021variable,
  title={Variable stiffness folding joints for haptic feedback},
  author={Zuliani, Fabio and Paik, Jamie},
  booktitle={2021 IEEE/RSJ International Conference on Intelligent Robots and Systems (IROS)},
  pages={8332--8338},
  year={2021},
  organization={IEEE}
}

@article{mcclintock2018millidelta,
  title={The milliDelta: A high-bandwidth, high-precision, millimeter-scale Delta robot},
  author={McClintock, Hayley and Temel, Fatma Zeynep and Doshi, Neel and Koh, Je-sung and Wood, Robert J},
  journal={Science Robotics},
  volume={3},
  number={14},
  pages={eaar3018},
  year={2018},
  publisher={American Association for the Advancement of Science}
}

@article{mintchev2019portable,
  title={A portable three-degrees-of-freedom force feedback origami robot for human--robot interactions},
  author={Mintchev, Stefano and Salerno, Marco and Cherpillod, Alexandre and Scaduto, Simone and Paik, Jamie},
  journal={Nature Machine Intelligence},
  volume={1},
  number={12},
  pages={584--593},
  year={2019},
  publisher={Nature Publishing Group UK London}
}

@article{williams20224,
  title={A 4-degree-of-freedom parallel origami haptic device for normal, shear, and torsion feedback},
  author={Williams, Sophia R and Suchoski, Jacob M and Chua, Zonghe and Okamura, Allison M},
  journal={IEEE Robotics and Automation Letters},
  volume={7},
  number={2},
  pages={3310--3317},
  year={2022},
  publisher={IEEE}
}

@article{choi2023fabrication,
  title={Fabrication of Origami Soft Gripper Using On-Fabric 3D Printing},
  author={Choi, Hana and Park, Tongil and Hwang, Gyomin and Ko, Youngji and Lee, Dohun and Lee, Taeksu and Park, Jong-Oh and Bang, Doyeon},
  journal={Robotics},
  volume={12},
  number={6},
  pages={150},
  year={2023},
  publisher={MDPI}
}

@article{faber2018bioinspired,
  title={Bioinspired spring origami},
  author={Faber, Jakob A and Arrieta, Andres F and Studart, Andr{\'e} R},
  journal={Science},
  volume={359},
  number={6382},
  pages={1386--1391},
  year={2018},
  publisher={American Association for the Advancement of Science}
}

@inproceedings{zhakypov2019programmable,
  title={Programmable fluidic networks design for robotic origami sequential self-folding},
  author={Zhakypov, Zhenishbek and Mete, Mustafa and Fiorentino, Julien and Paik, Jamie},
  booktitle={2019 2nd IEEE International Conference on Soft Robotics (RoboSoft)},
  pages={814--820},
  year={2019},
  organization={IEEE}
}

@article{kim2019bioinspired,
  title={Bioinspired dual-morphing stretchable origami},
  author={Kim, Woongbae and Byun, Junghwan and Kim, Jae-Kyeong and Choi, Woo-Young and Jakobsen, Kirsten and Jakobsen, Joachim and Lee, Dae-Young and Cho, Kyu-Jin},
  journal={Science robotics},
  volume={4},
  number={36},
  pages={eaay3493},
  year={2019},
  publisher={American Association for the Advancement of Science}
}

@article{lee2021high,
  title={High--load capacity origami transformable wheel},
  author={Lee, Dae-Young and Kim, Jae-Kyeong and Sohn, Chang-Young and Heo, Jeong-Mu and Cho, Kyu-Jin},
  journal={Science Robotics},
  volume={6},
  number={53},
  pages={eabe0201},
  year={2021},
  publisher={American Association for the Advancement of Science}
}

@inproceedings{russo2016soft,
  title={Soft pop-up mechanisms for micro surgical tools: Design and characterization of compliant millimeter-scale articulated structures},
  author={Russo, Sheila and Ranzani, Tommaso and Gafford, J and Walsh, Conor J and Wood, Robert J},
  booktitle={2016 IEEE International Conference on Robotics and Automation (ICRA)},
  pages={750--757},
  year={2016},
  organization={IEEE}
}

@book{craig2020mechanics,
  title={Mechanics of materials},
  author={Craig Jr, Roy R and Taleff, Eric M},
  year={2020},
  publisher={John Wiley \& Sons}
}

%




%
\end{document}